\documentclass[10pt,twocolumn,letterpaper]{article}

\usepackage{iccv}
\usepackage{times}
\usepackage{epsfig}
\usepackage{graphicx}
\usepackage{amsmath}
\usepackage{amssymb}

\usepackage{url}
\usepackage{booktabs}
\usepackage{microtype}
\usepackage[table]{xcolor}
\usepackage{multirow}

\usepackage[pagebackref=true,breaklinks=true,letterpaper=true,colorlinks,bookmarks=false]{hyperref}

\iccvfinalcopy 

\def\iccvPaperID{****} 

\ificcvfinal\fi
\begin{document}

\title{RethNet: Object-by-Object Learning for Detecting Facial Skin Problems}

\author{Shohrukh Bekmirzaev\\
lululab Inc\\
Gangnam-gu, Seoul, Korea\\
{\tt\small shoh.bek@lulu-lab.com}
\and
Seoyoung  Oh\\
lululab Inc\\
Gangnam-gu, Seoul, Korea\\
{\tt\small seoyoung.oh@lulu-lab.com} 
\and
Sangwook  Yoo\\
lululab Inc\\
Gangnam-gu, Seoul, Korea\\
{\tt\small sangwook.yoo@lulu-lab.com}
}
\maketitle
\thispagestyle{empty}

\begin{abstract}
Semantic segmentation is a hot topic in computer vision
where the most challenging tasks of object detection and
recognition have been handling by the success of semantic
segmentation approaches. We propose a concept of object-by-object learning technique to detect 11 types of facial
skin lesions using semantic segmentation methods. Detecting individual skin lesion in a dense group is a challenging
task, because of ambiguities in the appearance of the visual data. We observe that there exist co-occurrent visual
relations between object classes (e.g., wrinkle and age spot,
or papule and whitehead, etc.). In fact, rich contextual
information significantly helps to handle the issue. Therefore, we propose REthinker blocks that are composed of
the locally constructed convLSTM/Conv3D layers and SE
module as a one-shot attention mechanism whose responsibility is to increase network’s sensitivity in the local and
global contextual representation that supports to capture ambiguously appeared objects and co-occurrence interactions
between object classes. Experiments show that our proposed
model reached MIoU of 79.46\% on the test of a prepared
dataset, representing a 15.34\% improvement over Deeplab
v3+ (MIoU of 64.12\%).

\end{abstract}


\section{Introduction}

\begin{figure}[t]
\begin{center}
\includegraphics[width=\linewidth]{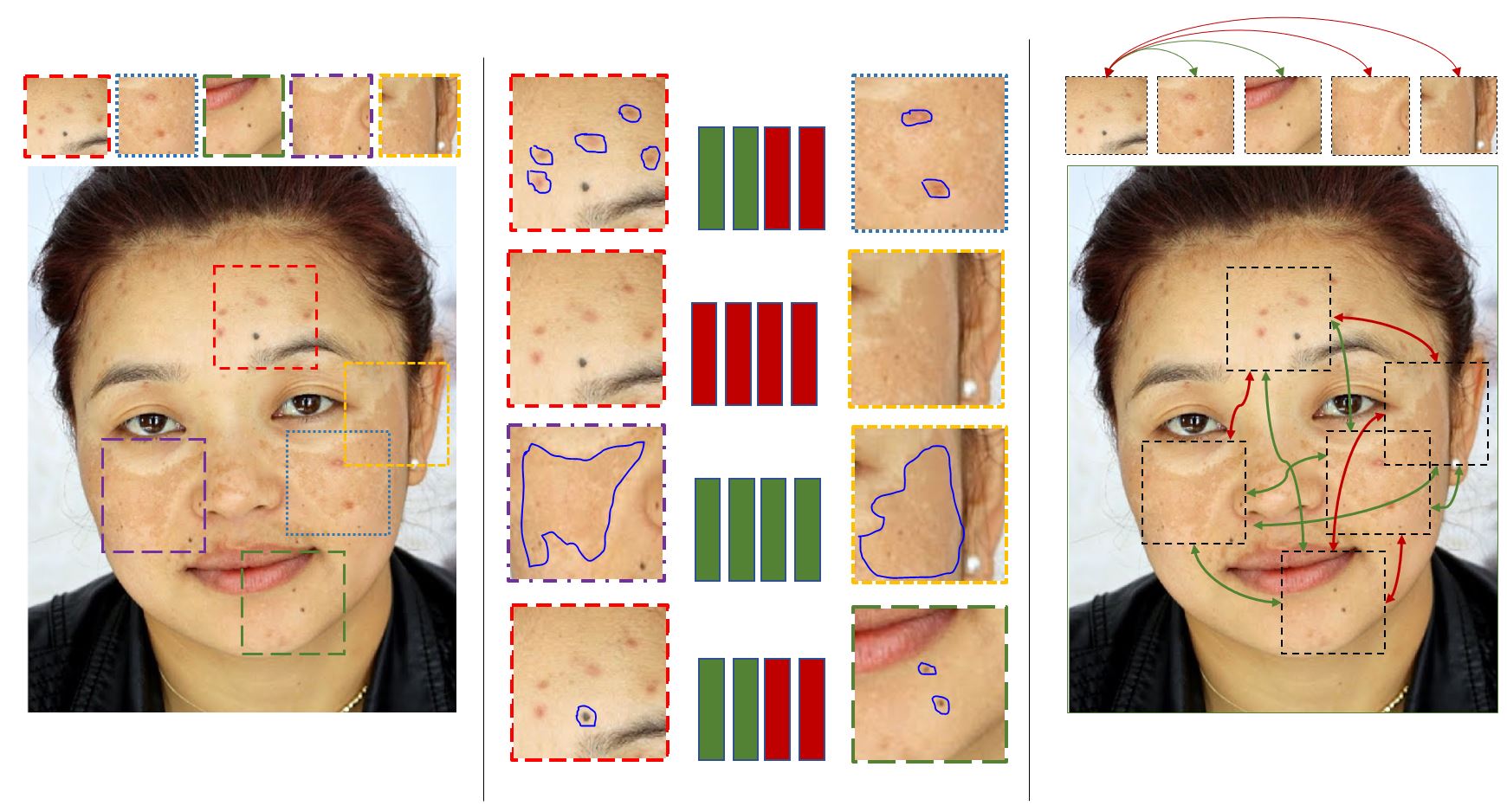}
\end{center}
\caption{ \textbf{ The concept of object-by-object learning}: In fact, skin lesion objects have visual relations between each other where it helps easily human to judge about what type of skin lesions are. In the figure, the green and red boxes represent the level of the image patch's similarity. The green and red line describe that there exist positive and negative relations between objects or a group of objects of each patch.
}
\label{fig:long}
\label{fig:sample}
\end{figure}

\begin{figure*}[t]
\begin{center}
\includegraphics[width=0.9\linewidth]{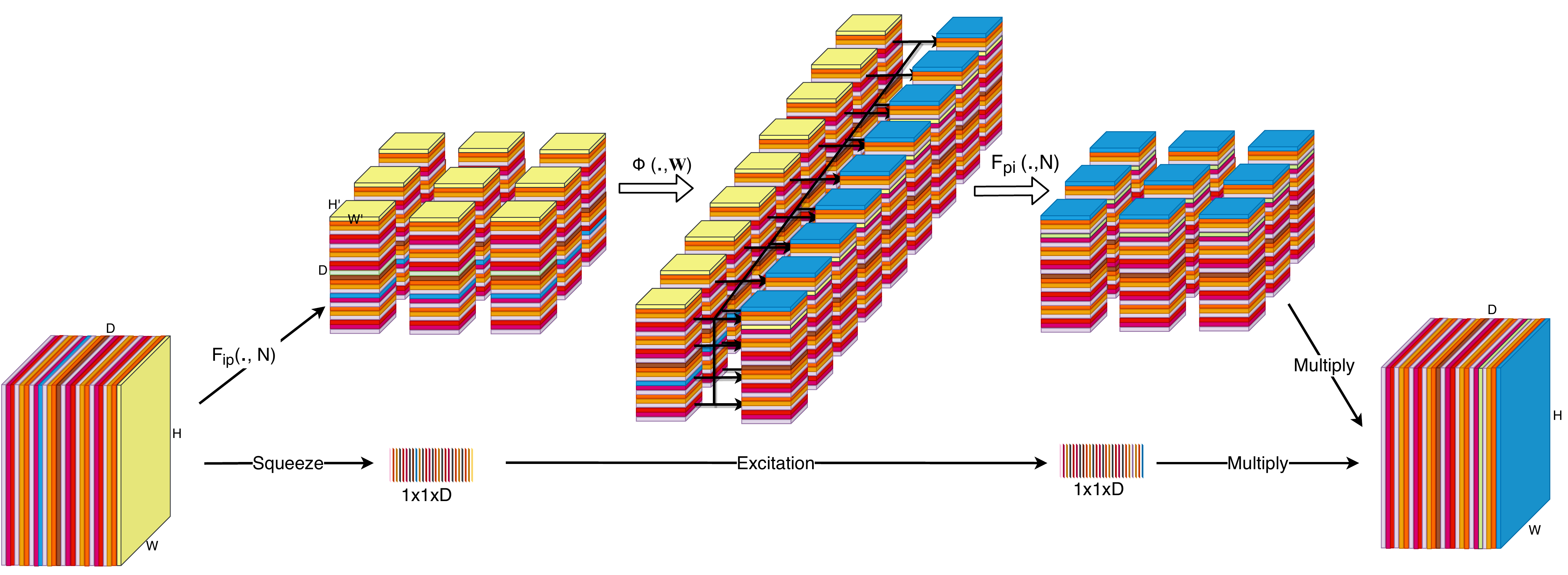}
\end{center}
\caption{ Proposed REthinker block that consists of the locally constructed convLSTM/Conv3D layers and SE module \cite{DBLP:journals/corr/abs-1709-01507} as a one-shot attention mechanism whose responsibility is to improve network’s sensitivity in local and global contextual representation that helps to capture ambiguously appeared objects and co-occurrence interactions between object classes.
}
\label{fig:long}
\label{fig:sample}
\end{figure*}

Semantic segmentation has been one of the fundamental and active topic in computer vision for a long time. This topic is of wide interest for real-world applications of autonomous driving, robotics and a range of medical imaging applications. Recent improvements and advances in semantic segmentation enable to emerge various new application areas in skin analysis. For example, facial skin lesion analysis has been attracting a lot of attention as having beautiful skin without troubles is getting popular and influenced on the society nowadays. E -cosmetics which involve the beautification, facial image simulation, digital makeup and accurate facial lesion analysis are fast-growing sector in the marketing.

Accurately and early detecting facial skin problems is an important clinical task and automated dermatology can be used to save time and reduce costs \cite{GI:1344-2014}. E-cosmetics and dermatological computer-aided systems are developing rapidly behind computer vision progresses \cite{article1825_1835}, \cite{KOROTKOV201269}, \cite{GI:1344-2014} , \cite{Tsumura:2003:ISC:882262.882344}, \cite{Esteva2017}, \cite{HAN20181529}, \cite{Liao2015ADL}. However, previous methods have shown only limited improvements and visual understanding of individual skin lesion in a dense group is still a challenging task.This is because, it is hard to distinguish some types of
facial skin lesions between each other as the fine-grained object categorization problem. Furthermore, facial skin lesions appear ambiguously with different (typically small) sizes, which lead a network to assign wrong classes easily. The use of rich contextual relation information helps to reduce the issue significantly \cite{Shotton2009}, \cite{1544868}. For example, there are the object-object interactions \cite{1544868} between some skin lesions where the detection of an object class helps to detect another by their co-occurrence interactions. The detection decisions about individual skin lesions can be switched dynamically through contextual relations among objects. We denote this cognitive process as \textit{object-by-object} decision-making.

We present a \textit{REthinker} module based on the SENet module \cite{DBLP:journals/corr/abs-1709-01507} \cite{DBLP:journals/corr/abs-1810-12348} and locally constructed convLSTM/conv3D unit \cite{DBLP:journals/corr/ShiCWYWW15} to increase network’s sensitivity in local and global contextual representations. The proposed modules are easy to use and applicable in any standard convolutional neural networks (CNNs). The use of the \textit{REthinker} modules forces networks to capture the contextual relationships between object classes regardless of similar texture and ambiguous appearance they have.

We experiment our proposed modules by modifying current state-o- the-art networks \cite{DBLP:journals/corr/HeZRS15}, \cite{DBLP:journals/corr/SzegedyIV16}, \cite{ DBLP:journals/corr/Chollet16a} for feature extraction. We originally use the decoder of DeepLabv3+ \cite{DBLP:journals/corr/abs-1802-02611}. Experimental results show that our proposed models outperform the state-of-the-art segmentation networks \cite{Ma_2018_ECCV},\cite{Yang_2018_CVPR}, \cite{DBLP:journals/corr/abs-1802-02611} ,\cite{DBLP:journals/corr/ChenPK0Y16}, \cite{DBLP:journals/corr/abs-1803-08904}, \cite{DBLP:journals/corr/abs-1801-04381} by the high difference of 15.34\% MIoU \cite{DBLP:journals/corr/ChenPSA17}, \cite{DBLP:journals/corr/HowardZCKWWAA17}, \cite{DBLP:journals/corr/ZhangZLS17} in detecting of facial skin problems in a prepared dataset. Moreover, our models have shown promising results on ISIC 2018 segmentation tasks.
The overall contributions of our paper can be summarized as follows:
\begin{itemize}
\item We introduce a new concept named "object-by-object" learning, where an object can be identified by looking at other objects.
\item We propose a novel residual block called REthinker modules that support "object-by-object" technique by capturing contextual relationships between object classes.
\item We develop a novel RethNet architecture that detects skin lesions with higher accuracy than the recent state-of-the-art segmentation approaches.
\end{itemize}


\section{Related Work}
\textbf{Semantic Scene Understanding}: Semantic understanding of visual scenes has become ubiquitous in computer vision. Impressive semantic segmentation approaches are mostly based on the Fully Convolutional Network (FCNs) \cite{DBLP:journals/corr/LongSD14}, \cite{DBLP:journals/corr/DaiH015}, \cite{DBLP:journals/corr/LinSRH15}, \cite{DBLP:journals/corr/LiuLLLT15}, \cite{Zheng:2015:CRF:2919332.2919659} ,\cite{DBLP:journals/corr/ChenPKMY14} , \cite{Noh:2015:LDN:2919332.2919658}. One key reason for the success of FCNs is that they use multi-scale (MS) image representations, which are subsequently upsampled to recover the lost resolution. Moreover, atrous convolution has proven to be an effective technique by providing a larger receptive field size without increasing the numbers of kernel parameters \cite{DBLP:journals/corr/ChenPKMY14}. Spatial pyramid pooling module \cite{DBLP:journals/corr/HeZR014} has been successfully applied with atrous convolutions by state-of-the-art networks \cite{DBLP:journals/corr/ChenPSA17}, \cite{Yang_2018_CVPR},\cite{DBLP:journals/corr/HowardZCKWWAA17}, \cite{DBLP:journals/corr/ZhangZLS17}, \cite{Ma_2018_ECCV}, \cite{DBLP:journals/corr/ChenPK0Y16}, \cite{DBLP:journals/corr/abs-1803-08904},\cite{DBLP:journals/corr/abs-1802-02611} , \cite{DBLP:journals/corr/abs-1801-04381} on segmentation and object detection benchmarks \cite{Everingham:2010:PVO:1747084.1747104}, \cite{Mottaghi_2014_CVPR}, \cite{Everingham15}, \cite{DBLP:journals/corr/LinMBHPRDZ14}, \cite{Geiger:2013:VMR:2528331.2528333}, \cite{DBLP:journals/corr/CordtsORREBFRS16}. Recently, depthwise separable convolution \cite{DBLP:journals/corr/Chollet16a} has known as an efficient technique to reduce the computation complexity in convolutional operations and allow networks to go deeper \cite{DBLP:journals/corr/ChenPSA17}, \cite{DBLP:journals/corr/abs-1801-04381}, \cite{DBLP:journals/corr/HowardZCKWWAA17}, \cite{DBLP:journals/corr/ZhangZLS17}, \cite{Ma_2018_ECCV}. Originally, the depthwise separable convolution consists of the \textit{depthwise} and \textit{ pointwise} convolutions where depthwise convolution keeps the channels separate and uses the standard convolution operation in each input channel \cite{DBLP:journals/corr/Chollet16a}.

\textbf{Contextual relations:} H.S. Hock \textit{et al.} \cite{Hock1974} introduced early the contextual relations between object scenes by experiments based on the criteria of familiarity, physical plausibility, and belongingness. The scene context information plays a crucial role on the semantic scene understanding. However, the contextual relation information between object classes are often ignored \cite{8264816}, \cite{Shotton2009}, \cite{1544868}. The relative spatial configurations of particular objects yield the higher-level contextual information while the lower-level contextual information demonstrates the semantic and visual relationships among objects or group of objects. S. Kumar \textit{et al.} \cite{1544868} categorize types of the contextual relationships of the scene labeling to \textit{region-region, object-region} and \textit{object-object} interactions and provide a hierarchical framework using Conditional Random Fields (CRFs) for semantic segmentation. CRFs models are widely used to capture the local contextual interactions of image regions \cite{article2004} , \cite{Shotton2009}, \cite{Mottaghi:2014:RCO:2679600.2680106}, \cite{DBLP:journals/corr/ChenPSA17}, \cite{ 8100033}. RNNs are investigated broadly to aggregate global context in the semantic segmentation \cite{DBLP:journals/corr/YanZJBY16}, \cite{7298977}, \cite{DBLP:journals/corr/abs-1712-00617}, \cite{8264816} \cite{DBLP:journals/corr/LiangSXFLY15}, \cite{8451830}. SE modules \cite{DBLP:journals/corr/abs-1709-01507} \cite{DBLP:journals/corr/abs-1810-12348} are successfully applied to capture global contextual information by simply exploiting CNN layers. \\



\section{Datasets}
We prepare a dataset called ”Multi-type Skin Lesion
Labelled Database” (MSLD) with pixel-wise labelling of
frontal face images. We report that the designing of MSLD
is unique in ML community where it is not available such
dataset with the labelling of multi-type skin lesions of facial
images. We further test the proposed models in the International Skin Imaging Collaboration (ISIC) dataset which
holds dermoscopic images with 5 types of skin problems. In
this section, we introduce the process of image accumulation
and annotation of MSLD dataset and announce about ISIC
dataset.

\begin{figure}[t]
\begin{center}
\includegraphics[width=\linewidth]{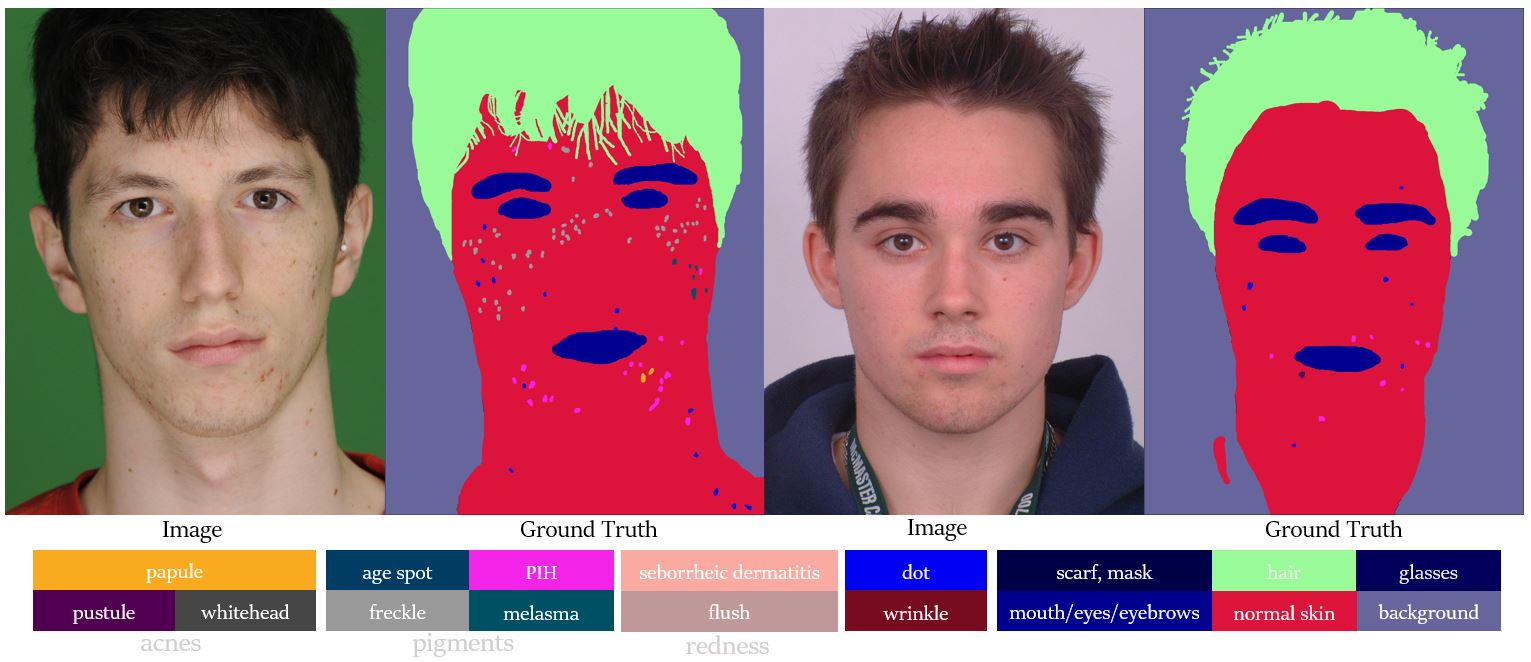}
\end{center}
\caption{Samples of the SiblingsDB  \cite{Vieira2014} dataset. Few samples
with skin problems have been annotated in order to show only the
labelling process of facial skin lesions. Note that we do not disclose
our MSLD dataset samples.}



\label{fig:long}
\label{fig:labelInfo}
\end{figure}

\subsection{Image accumulation}
We collected a total of 27,790 frontal face images using kiosks in cosmetics stores during the period from April to August in 2018. The kiosks have a standard camera whose image sensor is a 1/3.2 Inch CMOS IMX179. The user's consent is obtained before capturing images. The total number of pixels of the image sensor is 3,288 x 2,512 (8.26Mpixel) with 24-bit depth. The images are captured at autofocus and auto-white balance in the distance of 20 mm.

\subsection{Image Annotation}
The collected images are studied carefully. Then 412 images have been annotated with the labelling of 11 common types of facial skin lesions and 6 additional classes as in \autoref{fig:labelInfo} using the PixelAnnotation tool \cite{Breheret2017}. The skin lesions are \textit{whitehead, papule, pustule, freckle, melasma, age spots, PIH \footnote {Post inflammatory hyperpigmentation (PIH) caused usually by inflammation or acnes}, flush, seborrheic, dermatitis, wrinkle} and \textit{black dot}. The additional classes are normal skin, hair, eyes/mouth/eyebrow, glasses, mask/scarf and background. We report that we do not disclose our (MSLD) dataset as the user's privacy is taken under the responsibility. We use the collected and annotated images as research purposes where they are used only in the training.

\subsection{ISIC Dataset}
We test the proposed approaches in the TASK 2 \footnote {https://challenge2018.isic-archive.com/task2/} of the ISIC, 2018 challenge called "Lesion Attribute Detection" \cite{DBLP:journals/corr/abs-1710-05006} in order to provide further experiments. The goal of the task is to predict 5 skin lesion attributes from dermoscopic images. These lesion attributes are \textit{pigment networks, negative network, streaks, milia-like cysts, globules, and dots.} The lesion classes that have visual similarity can be seen as a fine-grained classification problem. There are 2594 images for training, 100 images for validation, and 1000 images for the test.

\begin{figure*}[t]
\begin{center}
\includegraphics[width=\linewidth]{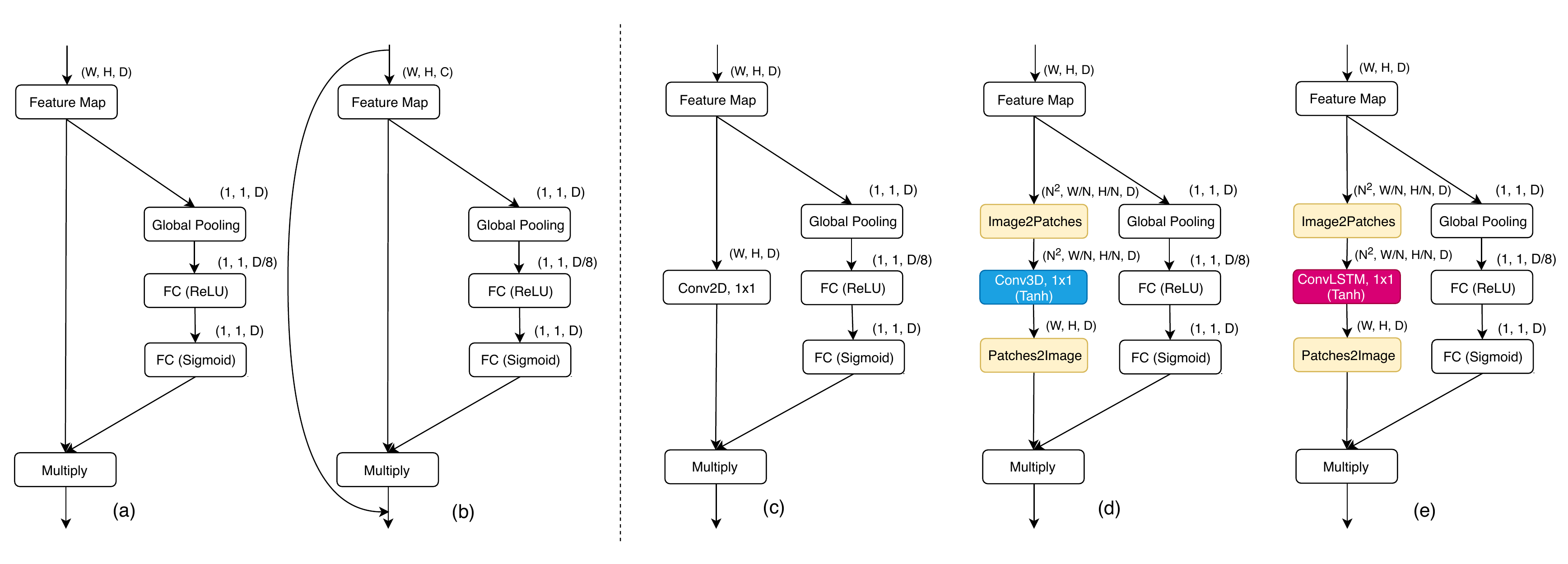}
\end{center}
\caption{ The figure represents comparison of SE blocks (a-b), baseline block (c) and REthinker blocks (d-e): The proposed REthinker blocks are designed under locally constructed Conv3D (d) and convLSTM (e) layers. }
\label{fig:ObByOb}
\end{figure*}

\section{Methods}

\subsection{ Squeeze and Excitation module }
Squeeze and Excitation network (SENet) \cite{DBLP:journals/corr/abs-1709-01507} has been introduced as a winner of the ILSVRC 2017 classification task in the top-5 error of 2.251\%. SE blocks \autoref{fig:ObByOb} [a-b] have proven to be an effective channel-wise attention mechanism \cite{DBLP:journals/corr/ChenZXNSC16} , which enables the network to perform dynamic channel-wise feature recalibration. SE module is computationally cheaper and helps the network to learn contextual higher-level features by the aggregated transformations of the global pooling. The SE block consists of squeeze and excitation operations where the \textit{squeeze} operation uses the global pooling to transform global spatial information to channel-wise statistics as a channel descriptor. The \textit{excitation} operation performs a self-gating mechanism based on 2 fully connected (FC) layers to capture channel-wise dependencies from the channel descriptor. Finally, the channel-wise dependencies are used to exploit the previous input transformation by the multiplication operation. The role of the SE module in our proposed module is to pass high-level contextual information outside of this region for context-dependent decision making.

\subsection{ REthinker module}
REthinker modules consist of SE module and the locally constructed convLSTM/conv3D layers as an one-shot attention mechanism as in \autoref{fig:ObByOb} [d-e], which are both responsible for extracting contextual relations from features. Precisely, as the global pooling of SE module aggregates the global spatial information, the SE module passes more embedded higher-level contextual information across large neighborhoods of each feature map. Whereas, the locally constructed convLSTM/conv3D layers encode lower-level contextual information across local neighborhoods elements of fragmented feature map (patches) while further take spatial correlation into consideration distributively over patches. The output of locally constructed convLSTM/conv3D recieves 3D , $ U_d \in R^{ H \times W \times D}$ feature maps from the residual blocks and passes a transformed 3D , $ U'_d \in R^{ H \times W \times D}$ feature map. The locally constructed convLSTM/conv3D is identified as follows:

\begin{equation}
\begin{aligned}
U'_d & = F_{pi} ( [\Phi (F_{ip} (U_d, N ) | v_t) | h_t], N)
\end{aligned}
\label{eq:localconv}
\end{equation}

Where, $F_{ip}$ is the \textit{ image2patches} operator function, $F_{ip} : R^{ H \times W \times C} \rightarrow R^{ N^2 \times H' \times W' \times D}$ to provide local spatiotemporal 4D data, $v_t = F_{ip} (U_d, N)$. Practically, the feature maps are transformed to patches over channel as spatiotemporal data. Here, $H' \times W' $ is a patch size $(H' = H / N, W'= W / N)$ to be assumed as an object or a group of objects. The given N is the dimensional slicing coefficient over the spatial dimensions (W, H) of the feature map. Thus, $\Phi$ is conv3D or convLSTM operator function, $ \Phi : R^{ N^2 \times H' \times W' \times D} \rightarrow R^{ N^2 \times H' \times W' \times D}$ to be applied with keeping the depth of the feature map. The convLSTM/conv3D serves to encode spatiotemporal correlations between features by veiwing sequentially objects or a group of objects of patches and passes the output as spatiotemporal data $h_t = \Phi (v_t)$ to the \textit{patches2image} operator whose identification function here is $F_{pi} : R^{ N^2 \times H' \times W' \times D} \rightarrow R^{ H \times W \times D}$ and $U'_d = F_{pi} (h_t, N)$. Note that $h_t = \Phi (v_t, h_{t-1})$ performs as the hidden states in the convLSTM. The gates $i_{t}, h_{t}, o_{t}$ of the convLSTM are identified as follows:

\begin{equation}
\begin{aligned}
i_{t} &= \sigma(w_{vi} \odot v_{t} + w_{hi} \odot h_{{t}-1} + w_{ci} \circ {c}_{{t}-1} + b_i) \\
f_{t} &= \sigma(w_{vf}\odot {v}_{t} + w_{hf}\odot {h}_{{t}-1} + w_{cf}\circ {c}_{{t}-1}+b_f) \\
{c}_{t} &= f_{t} \circ {c}_{t}-1 + i_{t} \circ \tanh(w_{vc} \ast {v}_{t} + w_{c} \odot {h}_{{t}-1}+b_c) \\
o_{t} &= \sigma(w_{vo}\odot v_{t} + w_{ho}\odot {h}_{{t}-1} + w_{co}\circ {c}_{t} +b_o) \\
{h}_{t} &= o_{t} \circ \tanh( {c}_t) 
\end{aligned}
\label{eq:localconv}
\end{equation}

Where, '$\odot$ ' represents the convolution operator and '$\circ$' denotes the Hadamard product. The output of SE module is used to exploit the output of locally constructed convLSTM/conv3D by the channel-wise multiplication operation as the output of REthinker module whose feature map represents long-range local and global contextual information to enable the context-dependent decision making.

\begin{figure*}[t]
\begin{center}
\includegraphics[width=\linewidth]{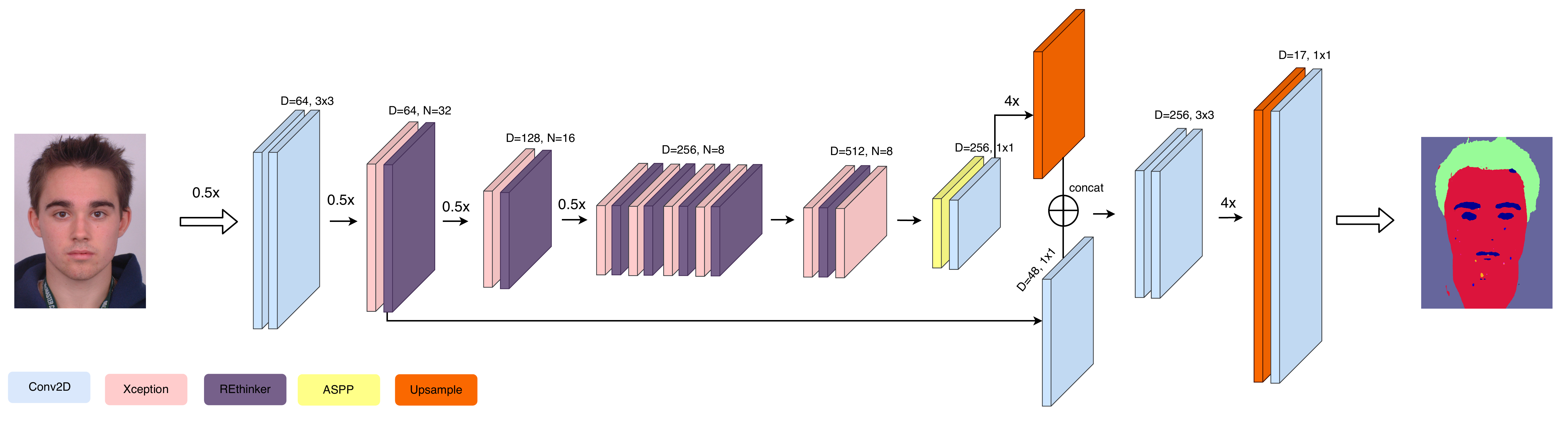}
\end{center}
\caption{Our proposed RethNet based RE-Xception and the decoder of the DeepLabv3+ \cite{DBLP:journals/corr/abs-1802-02611}. Where \textit{D} and \textit{N} is the depth of future maps and the dimensional slicing coefficient over the spatial dimensions respectively.}.
\label{fig:RethNet}
\end{figure*}

\subsection{ RethNeT }
\textbf {An Encoder Search:}
In practice, the REthinker blocks are applicable in any standard CNNs. We consider to employ current state of the art networks \cite{DBLP:journals/corr/HeZRS15}, \cite{DBLP:journals/corr/SzegedyIV16}, \cite{ DBLP:journals/corr/Chollet16a}.
We integrate the modern architectures with our proposed REthinker blocks to improve the network's sensitivity in local and global contextual representations enabling object-by-object learning technique. We experiment ResNet \cite{DBLP:journals/corr/HeZRS15}, ResNeXt \cite{DBLP:journals/corr/XieGDTH16} and Xception \cite{DBLP:journals/corr/Chollet16a}, \cite{DBLP:journals/corr/abs-1802-02611}.

\textbf {A Decoder Search:} We believe that the rich contextual information is a key to capture ambiguously appeared objects and co-occurrence
interactions between object classes where that is usually obtained in encoders. Therefore, we do not consider the decoder path. We select the decoder of DeepLabv3+ \cite{DBLP:journals/corr/abs-1802-02611} to recover object segmentation details of individual skin lesions.

\begin{table}
\footnotesize
\setlength{\tabcolsep}{2pt}
\begin{center}
\begin{tabular}{ l c c }
\hline
Models & Mean IoU(\%) & Pixel Acc(\%) \\
\hline\hline
DenseASPP \cite{Yang_2018_CVPR} & 58.31 & 89.31 \\
PSPNet+ResNet-101 \cite{DBLP:journals/corr/HeZR014} & 63.24 & 91.92 \\
DeepLabv3Plus + ResNet-101 \cite{DBLP:journals/corr/abs-1802-02611} & 63.74 & 92.51 \\
DeepLabv3Plus +ResNeXt-101 \cite{DBLP:journals/corr/LongSD14} & 64.64 & 93.14 \\
DeepLabv3Plus + Xception \cite{DBLP:journals/corr/abs-1802-02611} & 64.12 & 94.08 \\
\hline
DeepLabv3Plus + Xception+SE & 65.49 & 94.12 \\
DeepLabv3Plus + Xception+baseline-c & 65.52 & 94.21 \\
\hline
RethNet + baseline-c & 62.11 & 92.44 \\
RethNet + Rethinker-d & \textbf{76.56} & \textbf{96.45} \\
RethNet + Rethinker-e & \textbf{79.46} & \textbf{96.11} \\
\hline
\end{tabular}
\end{center}
\caption{ Experimental results on test samples of our MSLD dataset. }
\label{tab:baseline}
\end{table}

\textbf {RethNet:} We simply investigate RethNet with the combining of the Xception module and REthinker modules as in \autoref{fig:RethNet} . We modify Xception as follows: \textbf {(1)} We add REthinker module after each Xception blocks without spatial loss of feature maps. \textbf{(2)} We remove the final block of the \textit{entry flow} of Xception. \textbf{(3)} We keep the patch size as 4x4 in each REthinker module in order to "see" future maps wider in ConvLSTM/conv3D with simply increasing time steps. \textbf{(4)} The number of parameters is minimized in the \textit{ middle flow} and \textit{exit flow}. As suggested in \cite{DBLP:journals/corr/abs-1802-02611}, \textbf{(5)} the max-pooling operation is replaced by depthwise separable convolutions with striding and the batch normalization and ReLU is applied after each 3 x 3 depthwise convolution of the Xception module.


\begin{figure}[t]
\begin{center}
\includegraphics[width=0.75\linewidth]{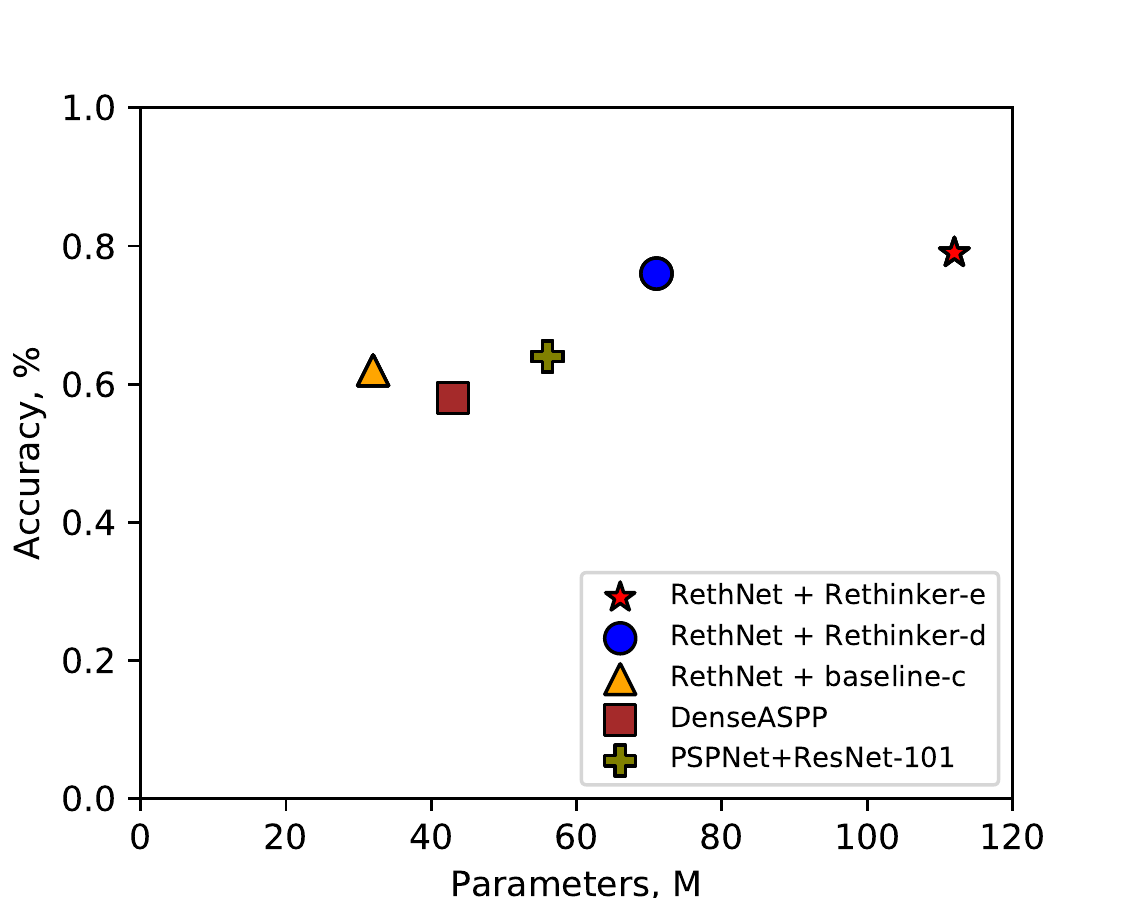}
\end{center}
\caption{ The number of parameters in the reference models and proposed modes with accuracy comparison on the test of MSLD dataset.
}
\label{fig:par}
\end{figure}

\begin{figure*}
\begin{center}
\includegraphics[width=0.85\linewidth]{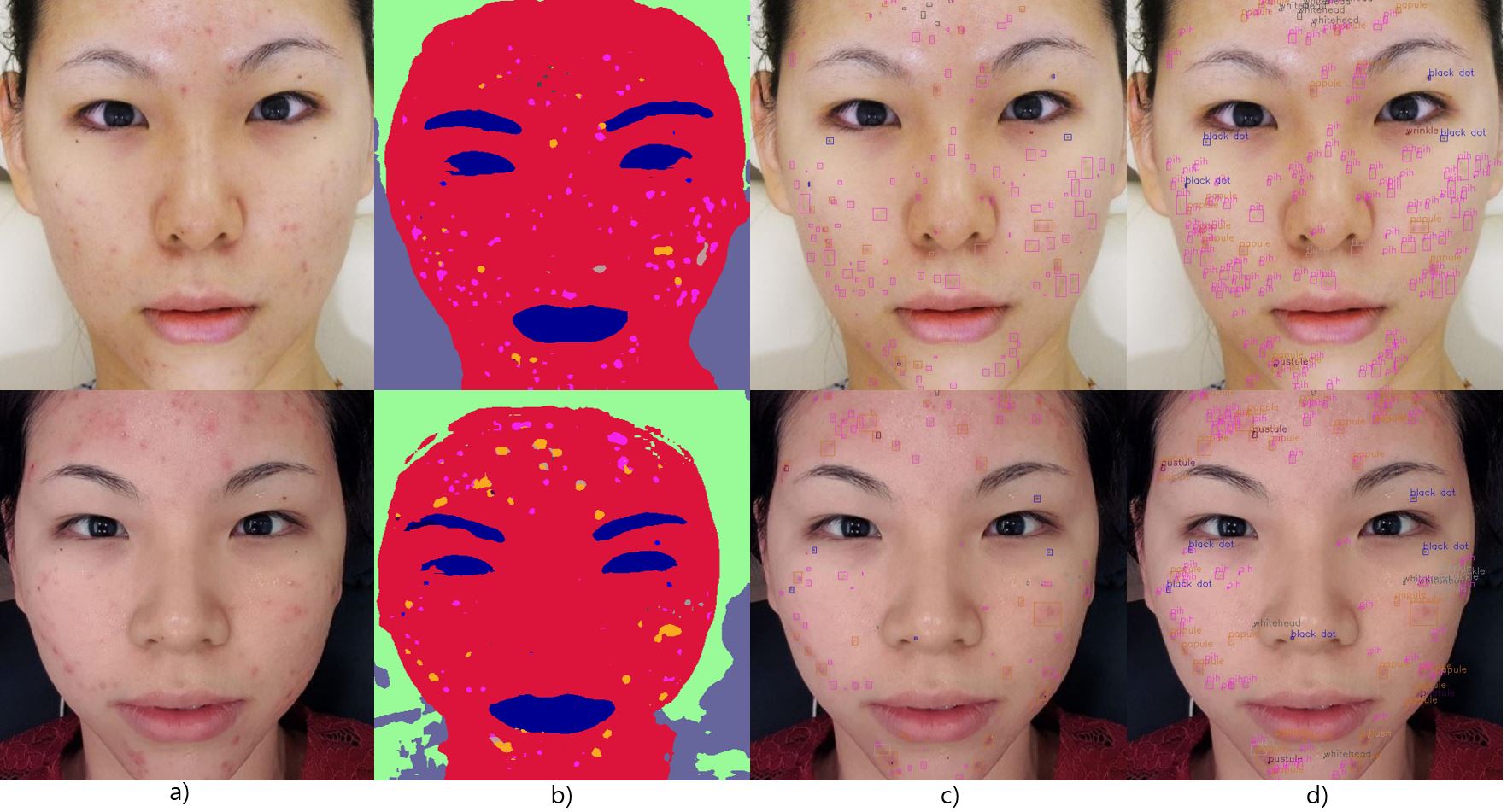}
\end{center}
\caption{The visualization results of \textbf{RehtNet + Rethinker-e:} The real test images (a) are obtained from the \href {https://fionaseah.com/category/beauty-2/health-aesthetics/face/}{ fionaseah.com}} with an agreement of the author.  The candidate of the test images is suffered from basically, PIH (caused by acne) and papules where the results images (b-d) show that the most of facial skin lesions are correctly predicted. 
\label{fig:short}
\end{figure*}

\section{Experiments}

\subsection{Implementation}

For experimental comparisons, we use the standard networks that are DeepLabv3+, PSPNet, DenseASPP as the reference networks. All models of networks were implemented using the TensorFlow framework and trained on a single NVIDIA GeForce GTX 1080 Ti GPUs, Intel(R) Core(TM) i7-8700K CPU @ 3.20GHz. In the experiments, 374 images are used for training and 38 images are for testing. We use the standard data augmentation techniques whose participants are the random rotation, random zooming, and random horizontal flipping during the training. The input and ground-truth images of the dataset is resized to 730 x 960 and random cropping applied by 512 x 512 as the input of the network during the training. We follow the same training protocols as suggested in \cite{DBLP:journals/corr/abs-1802-02611}, \cite{DBLP:journals/corr/HeZR014}. In all experiments, the softmax cross-entropy is applied for loss and the momentum optimizer is used, whose base learning rate is set to 0.001 decreased by a factor of 10 every 50 epoch of total 200 epoch with decay 0.9.

\begin{table}
\footnotesize
\setlength{\tabcolsep}{2pt}
\begin{center}
\begin{tabular}{ l c c c }
\hline
Models & Rank & Jaccard (\%) & Dice (\%) \\
\hline\hline

Ensemble \cite{DBLP:journals/corr/abs-1809-10243} & 1 & \textbf{0.483} & \textbf{ 0.651} \\
Unet + ASPP +DenseNet169 \cite{DBLP:journals/corr/abs-1809-10243} & 2 & 0.464 & 0.629 \\
Unet + ASPP + Resnetv2 \cite{DBLP:journals/corr/abs-1809-10243} & 3 & 0.455 & 0.616 \\
Unet + ASPP +ResNet151 \cite{DBLP:journals/corr/abs-1809-10243} & 5 & 0.436 & 0.598 \\
DeepLabv3Plus + Xception \ & - & 0.451& 0.614 \\
\hline
DeepLabv3Plus + Xception+SE & - & 0.469 & 0.627 \\
DeepLabv3Plus + Xception+baseline-c & - & 0.456 & 0.616 \\
\hline
RethNet + baseline-c & - & 0.441 & 0.592 \\
RethNet + Rethinker-d & - & 0.473 & 0.639 \\
RethNet + Rethinker-e & - & \textbf{0.475} & \textbf{0.644} \\
\hline
\end{tabular}
\end{center}
\caption{ Experimental results on test samples of ISIC 2018 challenge in the task 2. The evaluation
metrics in this task are average Jaccard Index (mIoU) and Dice coefficient following by the proposed metrics of the challenge. The rank represents positions on the test leaderboard of the challenge \cite{DBLP:journals/corr/abs-1809-10243}. Note that the Jaccard index metric, also referred to as the Intersection over Union (IoU). }
\label{tab:asic}
\end{table}
\subsection{Results}

Thanks to REthinker blocks, It shows significant improvements in the facial skin lesion detection task of MSLD dataset. As \autoref{tab:baseline} summarized results, we made a further improvement by DeepLabv3Plus + Xception+ SE and DeepLabv3Plus + Xception+ baseline-c, which showed a better result of 65.49 and 65.52 MIoU than all other reference models except the proposed models. Our proposed network RethNet+ REthinker-e blocks achieved MIoU of 79.46\% on the test, where it is the top in all experiments.

\begin{figure*}[t]
\begin{center}
\includegraphics[width=\linewidth]{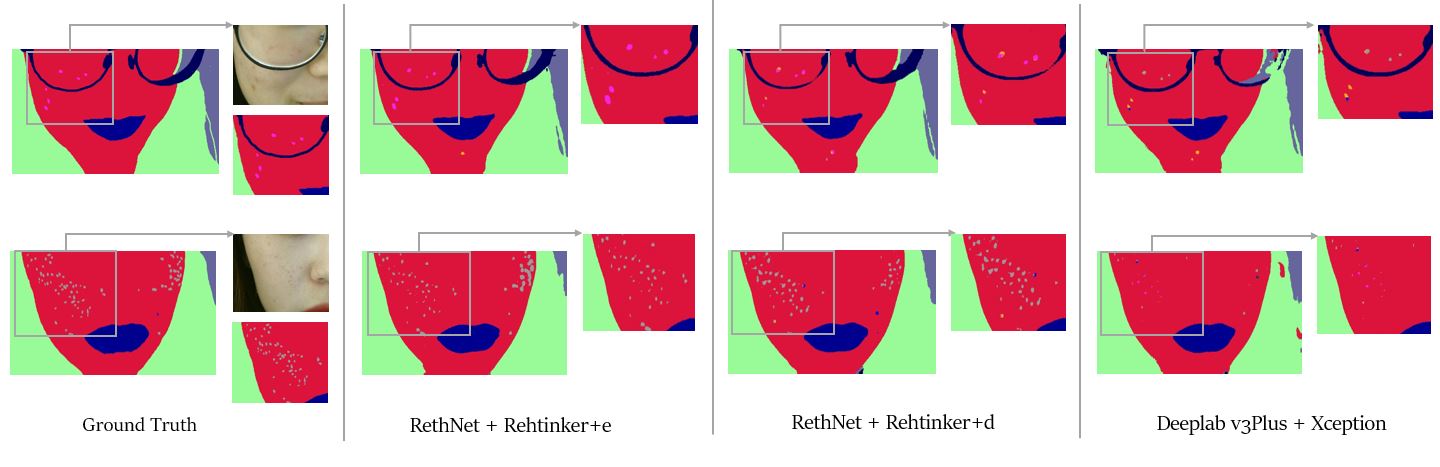}
\end{center}
\caption{The ground truth and inference results of proposed models and the best reference model in MSLD dataset . We keep the face entity of the MSLD dataset.}
\label{fig:long}
\label{fig:sample1}
\end{figure*}


\textbf{Computational cost:} We compare proposed models with the reference models in terms of the number of parameters and accuracy (\autoref{fig:par}). Even though our RethNet + REthinker-e block is the top on the list with 112M parameters, it reached high performance in the accuracy with a big difference (e.g 15 \% of mIoU greater than the best reference model, 14\% of mIoU higher than the baseline model). We further report inference time of the RethNet + REthinker-e block that whose running time for per image inference by the 512x512 of resolution is an average 2.7 sec on a single GPU and 11 sec on CPU of those mentioned hardware sources.

\subsection{ISIC challenge}
As \autoref{tab:asic} represents, our RethNet + REthinker-e block showed 47.5\% of Mean-Jaccard on the test of ISIC dataset, where it outperforms all competitive single models except an ensemble model ("48.3\% of Mean-Jaccard"). Note that the top-ranked models on the test leaderboards of the challenge use broadly preprocessing techniques such as image enhancing, polluting dermoscopic images with random hairs, and data augmentation. However, we apply only common techniques of data augmentation during training and test images are evaluated without any preprocessing techniques. \\

\subsection{Contribution Discussion} In fact, the fine-grained classification has been becoming an open issue in the computer vision community so far. It is a real challenge to differentiate classes that have similar visual context. Especially the problem is more common in medical imaging applications. We consider solving the problem with a novel straightforward technique by our application ("Detecting multi-type facial skin lesions") where there is not yet accurate and solid application or method to detect correctly and differentiate multi-type facial skin lesions. The proposed blocks are easy to use and possible to apply to any standard CNNs with considering time complexity by limiting the number of the blocks.


\section{Conclusion}
We propose successfully an efficient network architecture to address detecting multi-type facial skin lesions by a novel \textit{object-by-object} learning technique. Experimental results show that our proposed model outperformed state-of-the-art segmentation networks by a high gap in the MSLD dataset. Furthermore, our model takes promising results on the ISIC 2018 segmentation task. In the future, we consider the time complexity of Rethinker blocks and try to design more light-weight models.

\section{Acknowledgement} This work was supported and funded by SBA (Seoul Business Agency) under a contract number CY190018.

{\small
\bibliographystyle{ieee}
\bibliography{egbib}
}

\end{document}